\documentclass[final,5p,times,twocolumn]{elsarticle}
\usepackage{framed}
\usepackage{lineno}
\modulolinenumbers[5]

\journal{Medical Image Analysis}

\usepackage{hyperref}
\usepackage{eqnarray}
\usepackage{multirow}
\usepackage{pifont}
\usepackage{todonotes}
\usepackage{lscape}
\usepackage{subfig}
\usepackage{amsmath}
\usepackage{rotating}

\def\ie{\emph{i.e.}}
\def\eg{\emph{e.g.}}
\def\etc{\emph{etc}}
\def\etal{{\em et al.}}

\newcommand{\curl}[1]{\textcolor{red}{\url{#1}}}

\def\snm#1{{#1}}

\begin{document}
	

\begin{frontmatter}
\title{AGE Challenge: Angle Closure Glaucoma Evaluation in Anterior Segment Optical Coherence Tomography}

\author[2]{Huazhu~\snm{Fu}\fnref{my_note}\corref{cor1}}
\author[1]{Fei~\snm{Li}\fnref{my_note}\corref{cor1}}
\author[3]{Xu~\snm{Sun}\fnref{my_note}}
\author[3]{Xingxing~\snm{Cao}\fnref{my_note}}
\author[17]{Jingan~\snm{Liao}\fnref{my_note}}
\author[4,18]{Jos\'e~Ignacio~\snm{Orlando}\fnref{my_note}}

\author[14]{Xing~\snm{Tao}}
\author[5]{Yuexiang~\snm{Li}}

\author[13]{Shihao~\snm{Zhang}}
\author[13]{Mingkui~\snm{Tan}}

\author[15]{Chenglang~\snm{Yuan}}
\author[5]{Cheng~\snm{Bian}}

\author[10]{Ruitao~\snm{Xie}}
\author[10]{Jiongcheng~\snm{Li}}
 
\author[11]{Xiaomeng~\snm{Li}}
\author[11]{Jing~\snm{Wang}}
 
\author[6]{Le~\snm{Geng}}
\author[6]{Panming~\snm{Li}}

\author[7,8]{Huaying~\snm{Hao}}
\author[12,8]{Jiang~\snm{Liu}}

\author[9]{Yan~\snm{Kong}}
\author[9]{Yongyong~\snm{Ren}}

\author[16]{Hrvoje~\snm{Bogunovi\'c}\fnref{my_note}}
\author[1]{Xiulan~\snm{Zhang}\fnref{my_note}\corref{cor2}} 
\author[3]{Yanwu~\snm{Xu}\fnref{my_note}\corref{cor2}}
\author{for iChallenge-PACG study group\fnref{group}}

\address[1]{State Key Laboratory, Zhongshan Ophthalmic Center, Sun Yat-sen University, Guangzhou 510060, China.}
\address[2]{Inception Institute of Artificial Intelligence, Abu Dhabi, UAE.}
\address[3]{Intelligent Healthcare Unit, Baidu, Beijing, China.}
\address[17]{School of Computer Science and Engineering, South China University of Technology, Guangzhou, Guangdong, China.}
\address[4]{National Scientific and Technical Research Council, CONICET, Argentina.}
\address[18]{Yatiris Group, PLADEMA Institute, Universidad Nacional del Centro de la Provincia de Buenos Aires (UNICEN), Tandil, Argentina.}
\address[14]{School of Computer Science and Technology, Hangzhou Dianzi University, Hangzhou, China.}
\address[5]{Tencent Jarvis Lab, Shenzhen, China.}
\address[13]{School of Software Engineering, South China University of Technology, Guangzhou, China.}
\address[15]{School of Biomedical Engineering, Health Science Center, Shenzhen University, Shenzhen, China.}
\address[10]{School of Electronic and Information Engineering, Shenzhen University, Shenzhen, China.}
\address[11]{Department of Computer Science and Engineering, The Chinese University of Hong Kong, China.}
\address[6]{School of Electronic and Information Engineering, Soochow University, Suzhou, China.}
\address[7]{Ningbo University, Zhejiang, China}
\address[8]{Ningbo Institute of Industrial Technology, Chinese Academy of Sciences, Zhejiang, China}
\address[12]{Southern University of Science and Technology, Shenzhen, China}
\address[9]{Shanghai Jiaotong University, Shanghai, China.}
\address[16]{Laboratory for Ophthalmic Image Analysis, Department of Ophthalmology, Medical University of Vienna, Vienna, Austria.}

\cortext[cor1]{These authors contributed equally to the work.}
\cortext[cor2]{Corresponding authors: Yanwu Xu (ywxu@ieee.org), and Xiulan Zhang (zhangxl2@mail.sysu.edu.cn).}
\fntext[my_note]{These authors co-organized the AGE challenge. All others contributed results of their algorithms presented in the paper.}
\fntext[group]{iChallenge-PACG study group includes: 
	Yichi Zhang (Department of Ophthalmology, Sun Yat-sen Memorial Hospital, SunYat-sen University, Guangzhou, China),  
	Nuhui Li (Guangzhou aier eye hospital, Guangzhou, China),    
	Chunman Yang
(Department of Ophthalmology, The Second Affiliated Hospital of GuiZhou Medical University, Kaili, China),
	Huang Luo
(Department of Ophthalmology, The Affiliated Tranditional Chinese Medicine Hospital of Guangzhou Medical University, Guangzhou, China),
	Xingyi Li (Zhongshan Ophthalmic Center, Sun Yat-sen University, Guangzhou, China),
	Feiyan Deng
(Department of Ophthalmology, The Tranditional Chinese Medicine Hospital Of Guangdong Province, Guangzhou, China),
	Yi Sun (Zhongshan Ophthalmic Center, Sun Yat-sen University, Guangzhou, China),
	Rouxi Zhou (Zhongshan Ophthalmic Center, Sun Yat-sen University, Guangzhou, China).}


\begin{abstract}

Angle closure glaucoma (ACG) is a more aggressive disease than open-angle glaucoma, where the abnormal anatomical structures of the anterior chamber angle (ACA) may cause an elevated intraocular pressure and gradually lead to glaucomatous optic neuropathy and eventually to visual impairment and blindness. Anterior Segment Optical Coherence Tomography (AS-OCT) imaging provides a fast and contactless way to discriminate angle closure from open angle. Although many medical image analysis algorithms have been developed for glaucoma diagnosis, only a few studies have focused on AS-OCT imaging. In particular, there is no public AS-OCT dataset available for evaluating the existing methods in a uniform way, which limits progress in the development of automated techniques for angle closure detection and assessment. To address this, we organized the Angle closure Glaucoma Evaluation challenge (AGE), held in conjunction with MICCAI 2019. The AGE challenge consisted of two tasks: scleral spur localization and angle closure classification. For this challenge, we released a large dataset of 4800 annotated AS-OCT images from 199 patients, and also proposed an evaluation framework to benchmark and compare different models. During the AGE challenge, over 200 teams registered online, and more than 1100 results were submitted for online evaluation. Finally, eight teams participated in the onsite challenge. In this paper, we summarize these eight onsite challenge methods and analyze their corresponding results for the two tasks. We further discuss limitations and future directions. In the AGE challenge, the top-performing approach had an average Euclidean Distance of 10 pixels (10$\mu$m) in scleral spur localization, while in the task of angle closure classification, all the algorithms achieved satisfactory performances, with two best obtaining an accuracy rate of 100\%. These artificial intelligence techniques have the potential to promote new developments in AS-OCT image analysis and image-based angle closure glaucoma assessment in particular.

\end{abstract}


\end{frontmatter}

\section{Introduction}

\begin{figure}[!t]
	\centering
	\includegraphics[width = 1\linewidth]{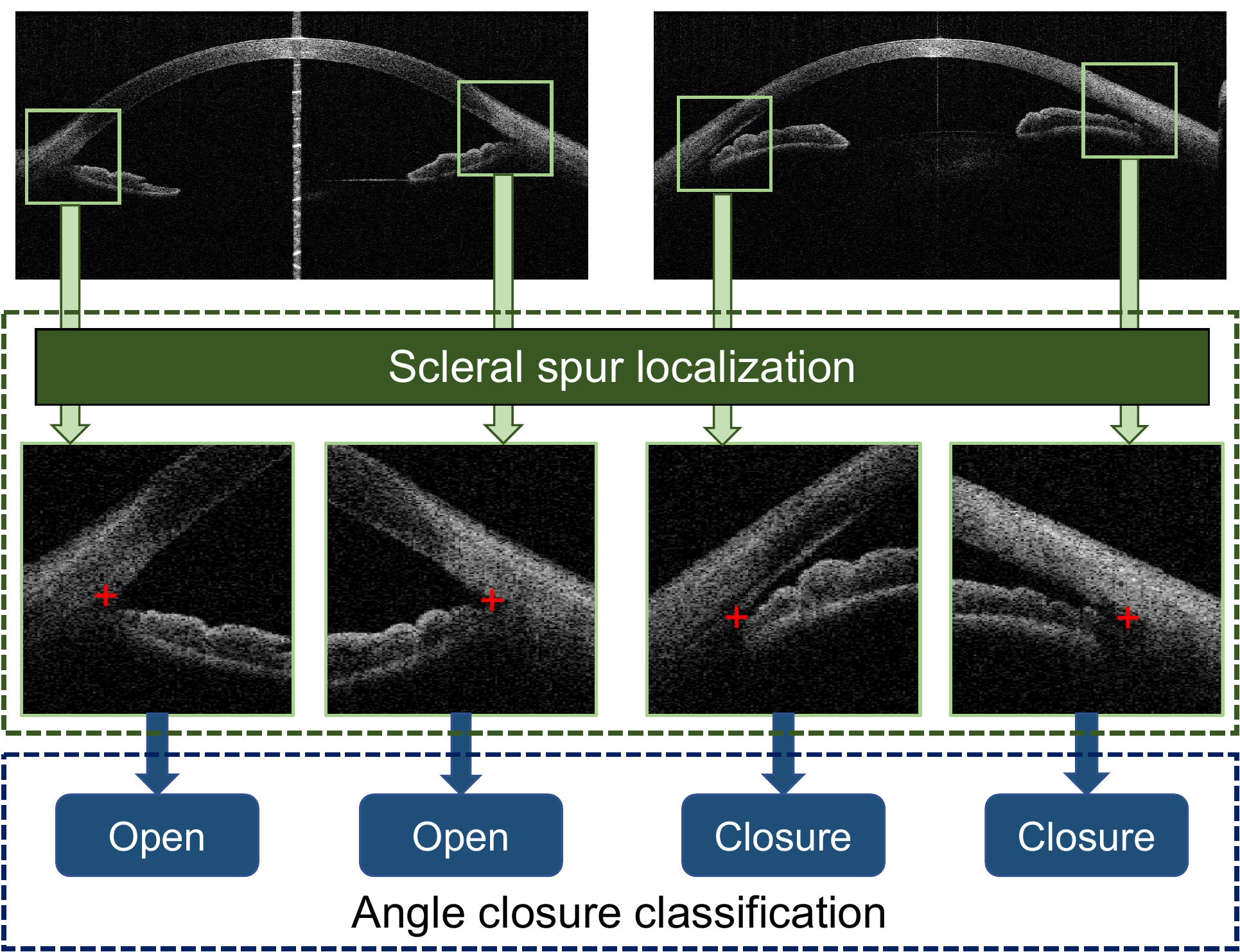}
	\caption{AGE challenge tasks: scleral spur localization and angle closure classification from AS-OCT images.}
	\label{fig-cover}
\end{figure}

As one of the world's main ocular diseases causing irreversible blindness, glaucoma involves both anterior and posterior segments of the eye. Primary angle closure glaucoma (PACG) is the main type of glaucoma in Asia~\citep{Quigley262,Foster1277,Chansangpetch2018}, where the abnormal anatomical structure of the anterior chamber angle (ACA) may cause elevated intraocular pressure and gradually lead to glaucomatous optic neuropathy. PACG patients have several characteristic structural differences from open-angle subjects~\citep{Nongpiur2013,Nongpiur2017}, including narrow chamber angles, short axial length, thick lens, greater iris thickness, etc. There are several ways to assess the angle structures for clinical diagnosis, \eg, gonioscopy, or anterior segment optical coherence tomography (AS-OCT). 
Gonioscopy is the current gold standard for the assessment and diagnosis of angle closure. Ophthalmologists grade the angle width into different levels according to the ACA structures seen under gonioscopy. However, being a contact examination, it may be uncomfortable for the patient, and it is also technically challenging, relying on the experience of the ophthalmologist in using this technique. 
By contrast, AS-OCT examination is a fast and contactless method for capturing the morphology of the ACA~\citep{Sharma2014,Ang2018}, which is easily used to identify open and narrow/closed angles. 
Moreover, AS-OCT imaging can obtain measurements of various angle parameters to assess the anterior chamber angle in a clinical setting~\citep{Sakata2008_Op,Nongpiur2017}, including angle open distance (AOD), anterior chamber width (ACW), trabecular iris space area (TISA), \etc. Quantification of these parameters relies on the localization of a specific mark, \ie, the scleral spur (SS), which appears as a wedge projecting from the inner aspect of the anterior sclera in cross-sectional images~\citep{Sakata2008}, as shown in Fig.~\ref{fig-cover}. Thus, SS localization is also a key task for identifying open and narrow/closed angles in AS-OCT imaging.  However, one limitation of AS-OCT imaging is that ACA assessment is time-consuming and subjective. For instance, the ophthalmologists have to manually identify specific anatomical structures, e.g., SS points, for detecting angle closure.


Recently, automated medical image analysis algorithms have achieved  promising performances in medicine and particularly ophthalmology~\citep{Schmidt-Erfurth2018,Ting2019_PRER,Rajkomar2019,Bi2019}. The availability of deep learning techniques has sparked tremendous global interest in major ophthalmic disease screening, including diabetic retinopathy (DR)~\citep{Gulshan2016,ShuWeiTing2017,Gargeya2017,Krause2018,Abramoff2018}, glaucoma~\citep{Asaoka2016,Li2018,MNet_2018,FuDeNet2018,REFUGE2020}, and age-related macular degeneration (AMD)~\citep{Grassmann2018,Kermany2018,DeFauw2018,DeepAMD2019,Peng2019}. 
However, most works focus on retinal fundus photographs, with only a few dealing with AS-OCT images~\citep{Niwas2016a,Fu2019_AJO,XU2019_AJO,Huaying2020,Jinkui2020}. Zhongshan Ophthalmic Center has provided a semi-automated angle assessment program to calculate various ACA parameters, but users are required to input the SS positions~\citep{zhongshan2008}. For fully automated systems, Tian \etal~\citep{Tian2011} provided a parameter calculation method for high-definition OCT (HD-OCT) based on the Schwalbe's line detection.
In~\citep{Fu2016EMBC,Fu2017ASOCT_TMI}, a label transfer system was proposed to combine segmentation, measurement and detection of AS-OCT structures. The major ACA parameters are recovered based on the segmented structure and serve as features for detecting anterior angle closure. 
Besides clinical parameter calculation, the visual features directly extracted from AS-OCT images using computer vision techniques are also utilized to classify angle-closure glaucoma. For instance, Xu \etal~\citep{Xu2012c,Xu2013} localized the ACA region and then extracted visual features to detect the glaucoma subtype. With the development of deep learning, Convolutional Neural Networks (CNNs) have been introduced to improve the performance of angle-closure detection in AS-OCT images~\citep{Fu2019_AJO,XU2019_AJO,8857615}. In~\citep{Fu2018_ASOCT_MICCAI,Fu2019_ASOCT_TC}, a multi-path deep network was designed to extract multi-scale AS-OCT representations for both the global image and clinically relevant local regions.
Nevertheless, these approaches cannot currently be properly compared due to the lack of a unified evaluation dataset. Moreover, the absence of a large-scale AS-OCT dataset also limits the rapid development and eventual deployment of deep learning techniques for angle closure detection.

To address these limitations, we introduced the Angle closure Glaucoma Evaluation (AGE) Challenge, a competition that was held as part of the Ophthalmic Medical Image Analysis (OMIA) workshop at the International Conference on Medical Image Computing and Computer Assisted Intervention (MICCAI) 2019. Our challenge followed on the success of the REFUGE challenge~\citep{REFUGE2020}, which was introduced for glaucoma detection in fundus image as part of iChallenge.
The challenge proposal was compliant with good MICCAI practices for biomedical challenges~\citep{Maier-Hein2018}.

The key contributions of the AGE challenge were: 
(1) The release of a large database of 4200 AS-OCT images with reliable reference standard annotations for SS localization and angle closure identification. \textbf{To the best of our knowledge, AGE was the first challenge to provide a public AS-OCT dataset for angle closure glaucoma.} 
(2) The construction of a unified evaluation framework that enables a standardized, fair comparison of different algorithms on \textbf{scleral spur localization and angle closure classification}, as shown in Fig.~\ref{fig-cover}. During the AGE challenge, more than 200 teams  registered online, and more than 1100 results were submitted for online evaluation. Eight teams participated in the final onsite challenge that took place in Shenzhen, China, during MICCAI 2019.   
In this paper, we analyze the outcomes and methodological contributions made as part of the AGE Challenge. We present and describe the competition and the released dataset, report the performance of the algorithms that participated in the onsite competition, and identify successful common practices for solving the tasks of the challenge. Finally, we take advantage of all this empirical evidence to discuss the clinical implications of the results and to propose further improvements to this evaluation framework. To encourage further developments and to ensure a proper and fair comparison of new proposals, AGE data and its associated evaluation platform remain open through the Grand Challenges website at~\curl{https://age.grand-challenge.org}.

\section{AGE Challenge Data}
 
The AS-OCT images used in the AGE Challenge were acquired with a CASIA SS-1000 OCT (Tomey, Nagoya, Japan) by the Zhongshan Ophthalmic Center, Sun Yat-sen University, China. The examinations were performed in a standardized darkroom with a light intensity lower than $0.4$ lux. Both left and right eyes of each patient were included if the images were eligible. Each AS-OCT volume contained 128 two-dimensional cross-sectional AS-OCT images (B-scans), which divided the anterior chamber into 128 meridians. 
We extracted 16 images from each volume equidistantly. Eyes with corrupt images or images with significant  eyelid artifacts precluding visualization of the ACA were excluded from the analysis. Angle structures were classified into open and closure. Gonioscopy was used as the gold standard. It was performed by a glaucoma expert (Zhang XL) with a four-mirror Sussman gonioscope (Ocular Instruments, Inc., Bellevue, WA) under standard dark illumination. The angle was graded in each quadrant (inferior, superior, nasal, and temporal) according to the modified Scheie classification system~\citep{SCHEIE1957} based on the identification of anatomical landmarks: grade 0, no structures visible; grade 1, non-pigmented trabecular meshwork (TM) visible; grade 2, pigmented TM visible; grade 3, SS visible; grade 4, ciliary body visible. A closed angle was diagnosed if the posterior  TM  was not seen for at least 180 degrees during static gonioscopy.

Each AS-OCT image was divided in two chamber angle images along the vertical middle-line. No adjustments were made to image brightness or contrast. For each chamber angle image, the SS was marked by four ophthalmologists (average experience: 8 years, range: 5-10 years) independently. The final standard reference SS localization was determined by the mean of these four independent annotations, followed by a fine adjustment by a senior glaucoma specialist (F.~Li). 
The study included 300 eyes from 199 subjects (female: 38.7\%, mean age: $47.2 \pm 15.4$). Each volume was composed of 128 radial images (B-scans). Adjacent images were similar to each other in chamber angle morphology. Therefore, we extracted only 16 images from each volume to avoid the influence of this similarity.  Thus, a total of 4800 images were extracted. Each image was composed of two chamber angle images, \eg, left and right. Finally, the dataset was split into a training set (1600 images with 640 angle closure and 2560 open angle), a validation set (1600 images with 640 angle closures and 2560 open angles), and a testing set (1600 images with 640 angle closures and 2560 open angles). Images from the same patient were assigned to the same set. The training set was used to learn the algorithm parameters (offline training), the validation set was used to choose a model (online evaluation), and the testing set was used to evaluate the model performance (onsite evaluation).

\section{Challenge Evaluation}

The performance of each proposed algorithm for each of the challenge tasks was assessed using different standard evaluation metrics. Each of them is described as follows.

\subsection{Task 1: Scleral Spur Localization}

\begin{figure}[!t]
	\centering
	\includegraphics[width = 1\linewidth]{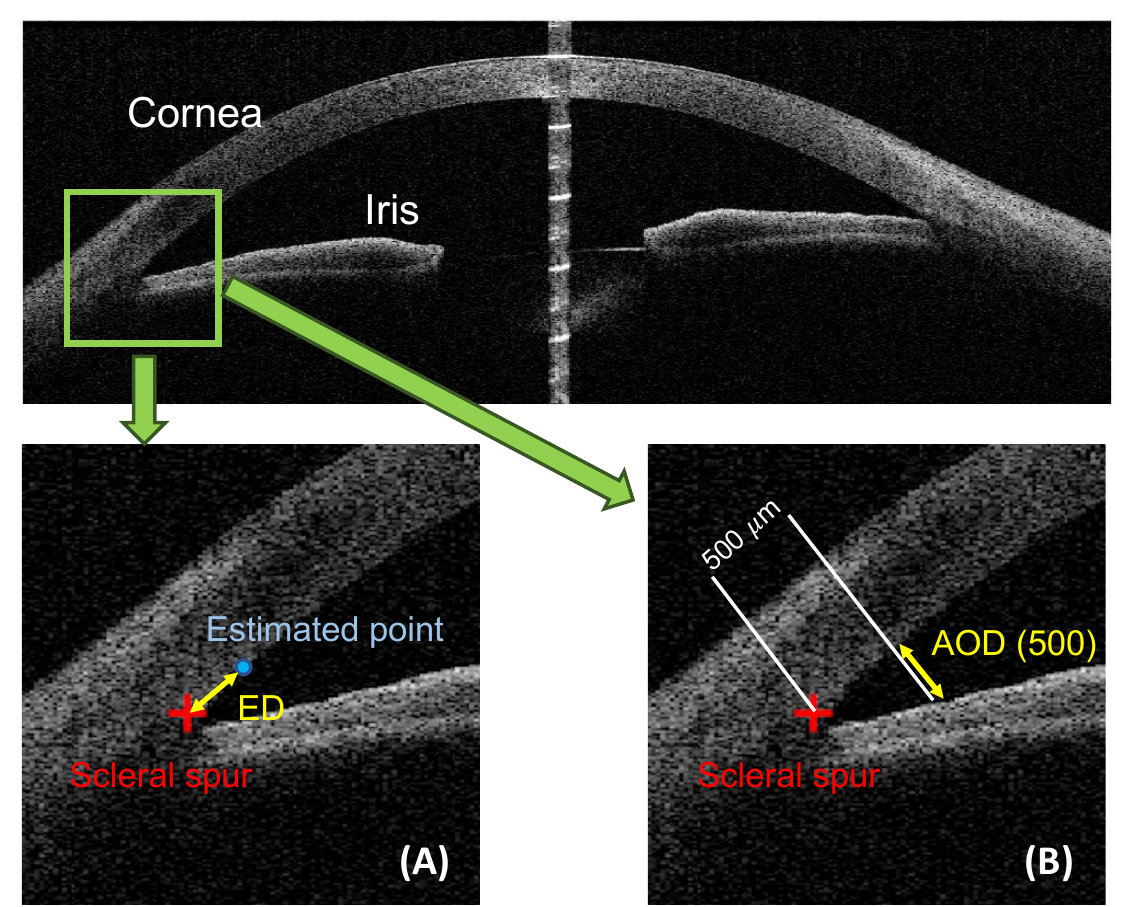}
	\caption{(A) Euclidean Distance ($ED$) measures the distance between the estimated point and ground truth of the scleral spur. (B) Angle Opening Distance (AOD) is the distance between the cornea and iris along a line perpendicular to the cornea at a specified distance (\eg, 500 $\mu$m).}
	\label{fig-SS_measure}
\end{figure}

Participants were asked to provide the estimated $(x, y)$ coordinates of the SS point. Submitted results were compared to the reference standard by means of two metrics. 
The first one was the Euclidean Distance ($ED$), which measures the distance between the estimated and ground truth SS locations, as shown in Fig.~\ref{fig-SS_measure} (A). 
The second criterion was the difference in the angle opening distance (AOD) ($\Delta_{AOD}$). AOD is defined as the distance between the cornea and iris along a line perpendicular to the cornea at a specified distance (in AGE, we used 500 $\mu$m) from the SS point~\citep{Chansangpetch2018}, as shown in Fig.~\ref{fig-SS_measure} (B). As an important indicator for angle closure assessment, we compared the AODs calculated using the prediction and ground truth. In general, the AOD of an open angle case is larger than that of angle closure. Thus, for an open angle image, we set a higher penalty for small calculated AODs, while for an angle closure image, we set a higher penalty for larger calculated AODs, as shown below:
\begin{itemize}
	\item For open angle images:
	\begin{equation}
		\Delta_{AOD} = \left\{
		\begin{aligned}
			\; 0.2 \times |z - z^*|, & & \text{if} \; z > z^*, \\
			\; 0.8 \times |z - z^*|, & & \text{otherwise}, \\
		\end{aligned}
		\right.
	\end{equation} 
	where $z$ and $z^*$ denote the AODs calculated using the estimated SS point and ground truth, respectively.
	\item For angle closure images:
	\begin{equation}
		\Delta_{AOD} = \left\{
		\begin{aligned}
			\; 0.8 \times |z - z^*|, & & \text{if} \; z > z^*, \\
			\; 0.2 \times |z - z^*|, & & \text{otherwise}. \\
		\end{aligned}
		\right.
	\end{equation} 
\end{itemize}
Based on the mean $ED$ and $\Delta_{AOD}$ values, each team received two ranks: $R_{ED}$ and $R_{AOD}$  (1 = best). The final ranking score for the scleral spur localization task was calculated as:
\begin{equation}
	S_{loc} = 0.4 \times R_{ED} + 0.6 \times R_{AOD},
\end{equation}
which was then used to determine the ranking of the SS localization leaderboard. We set a higher weight for $R_{AOD}$ because it could be used as an indicator for angle closure identification directly. Teams with lower ranking scores were ranked higher.

\subsection{Task 2: Angle Closure Classification}

Submissions for the classification challenge had to provide the corresponding estimated angle closure results (positive value for angle closure and non-positive value otherwise). Sensitivity and Specificity were utilized as the criteria of the challenge:
\begin{equation}
	\text{Sensitivity} = \frac{TP}{TP + FN} , \; \text{Specificity} = \frac{TN}{TN + FP} , 
\end{equation}
where $TP$ and $TN$ denote the number of true positives and true negatives, respectively, and $FP$ and $FN$ denote the number of false positives and false negatives, respectively. In addition, we also reported the area under receiver operating characteristic curve (AUC). 
Based on the Sensitivity, Specificity and AUC values, each team received three ranks: $R_{sen}$, $R_{spe}$ and $R_{AUC}$  (1 = best). The final ranking score for the angle closure classification task was calculated as:
\begin{equation}
	S_{cls} = 0.5 \times R_{AUC} + 0.25 \times R_{sen} + 0.25 \times R_{spe}.
\end{equation}

\subsection{Final Ranking}

The overall score of the onsite challenge was calculated as:
\begin{equation}
	S_{onsite} = 0.7  \times R_{loc} + 0.3  \times R_{cls},
	\label{eq_onsite_score}
\end{equation}
where $R_{loc}$ and $ R_{cls}$ denote the ranking scores of the SS localization and angle closure classification tasks, respectively.
A larger weight was set for the ranking of SS localization because the clinical measurements, \eg, AOD, derived from SS localization  can be used as a primary score for angle closure classification.

Eight teams  attended the final onsite challenge, which was held in Shenzhen, China, during  MICCAI 2019. The test set (only the images) was released during the workshop, and the eight teams had to submit their results within a time limit (three hours). The final submission of each team was taken into account for evaluation. Both online and onsite ranks were assigned to each team. The final rank of the challenge was based on a score $S_{final}$, calculated as the weighted average of the online and onsite rank positions:
\begin{equation}
	S_{final} = 0.3  \times S_{online} + 0.7  \times S_{onsite}.
	\label{eq_final_score}
\end{equation}
Note that a higher weight was assigned to the onsite results.
In this paper we only analyze the results from the onsite challenge, as it better reflects the generalization ability of the proposed solutions.

\begin{table*}[!t]
	\caption{A brief summary of the challenge methods on angle closure classification. CE = cross-entropy}
	\centering
	\footnotesize 
	\begin{tabular}{|p{40pt}|p{110pt}|p{60pt}|p{20pt}|p{100pt}|p{70pt}|}
		\hline
		Team      & Member                      & Architecture & ROI & Ensemble                & Loss                    \\ \hline\hline
		Cerostar  & Yan Kong, Yongyong Ren      & ResNet34     & No  & Single model            & CE loss                 \\ \hline
		CUEye     & Xiaomeng Li, Jing Wang      & SE-Net       & Yes & Three-scale ROIs        & CE loss                 \\ \hline
		Dream Sun & Chenglang Yuan,  Cheng Bian & ResNet152    & No  & Three trained model     & Focal loss, F-beta loss \\ \hline
		EFFUNET   & Xing Tao, Yuexiang Li       & EfficientNet & No  & EfficientNet B3, and B5 & CE loss                 \\ \hline
		iMed      & Huaying Hao, Jiang Liu      & ResNet50     & Yes & Three-scale ROIs        & CE loss                 \\ \hline
		MIPAV     & Le Geng,  Panming Li        & SE-ResNet18  & Yes & Single model            & Focal loss              \\ \hline
		Redscarf  & Shihao Zhang,  Mingkui Tan  & Res2Net      & Yes & Global and ROIs         & CE loss                 \\ \hline
		VistaLab  & Ruitao Xie,  Jiongcheng Li  & ResNet18     & Yes & Four-fold models        & CE loss                 \\ \hline
	\end{tabular} 
	\label{tab_team_classification}%
\end{table*}

\begin{table*}[!t]
	\caption{A brief summary of the challenge methods on scleral spur localization. MSE = mean squared error, CE = cross-entropy, ED = Euclidean Distance}
	\centering
	\footnotesize 
	\begin{tabular}{|p{40pt}|p{100pt}|p{20pt}|p{60pt}|p{110pt}|p{70pt}|}
		\hline
		Team      & Architecture               & ROI & Output            & Ensemble                        & Loss                 \\ \hline\hline
		Cerostar  & U-Net with ResNeXt34       & No  & Binary mask       & Four-scale                      & CE loss              \\ \hline
		CUEye     & Zoom-in SE-Net             & Yes & Value  regression & Three-scale ROIs                & MSE loss             \\ \hline
		Dream Sun & U-Net with EfficientNet    & Yes & Heat map          & EfficientNet B2, B3, B5, and B6 & MSE loss,  Dice loss \\ \hline
		EFFUNET   & U-Net with EfficientNet B5 & Yes & Heat map          & Single model                    & MSE loss             \\ \hline
		iMed      & GlobalNet,  ResNet34       & Yes & Value regression  & Single model                    & MSE loss             \\ \hline
		MIPAV     & LinkNet with ResNet18      & No  & Heat map          & Single model                    & MSE loss             \\ \hline
		Redscarf  & YOLO-V3, AG-Net            & Yes & Heat map          & Single model                      & MSE loss             \\ \hline
		VistaLab  & U-Net, VGG19               & Yes & Value regression  & Single model                    & ED loss              \\ \hline
	\end{tabular} 
	\label{tab_team_localization}%
\end{table*}

\section{Summary of Challenge Solutions}
\label{sec-method}

In the AGE challenge, we provided a unified evaluation framework for the standardized and fair comparison of different algorithms on two clinically relevant tasks: scleral spur localization and angle closure classification, as shown in Fig.~\ref{fig-cover}.

\begin{figure}[!t]
	\centering
	\includegraphics[width = 1\linewidth]{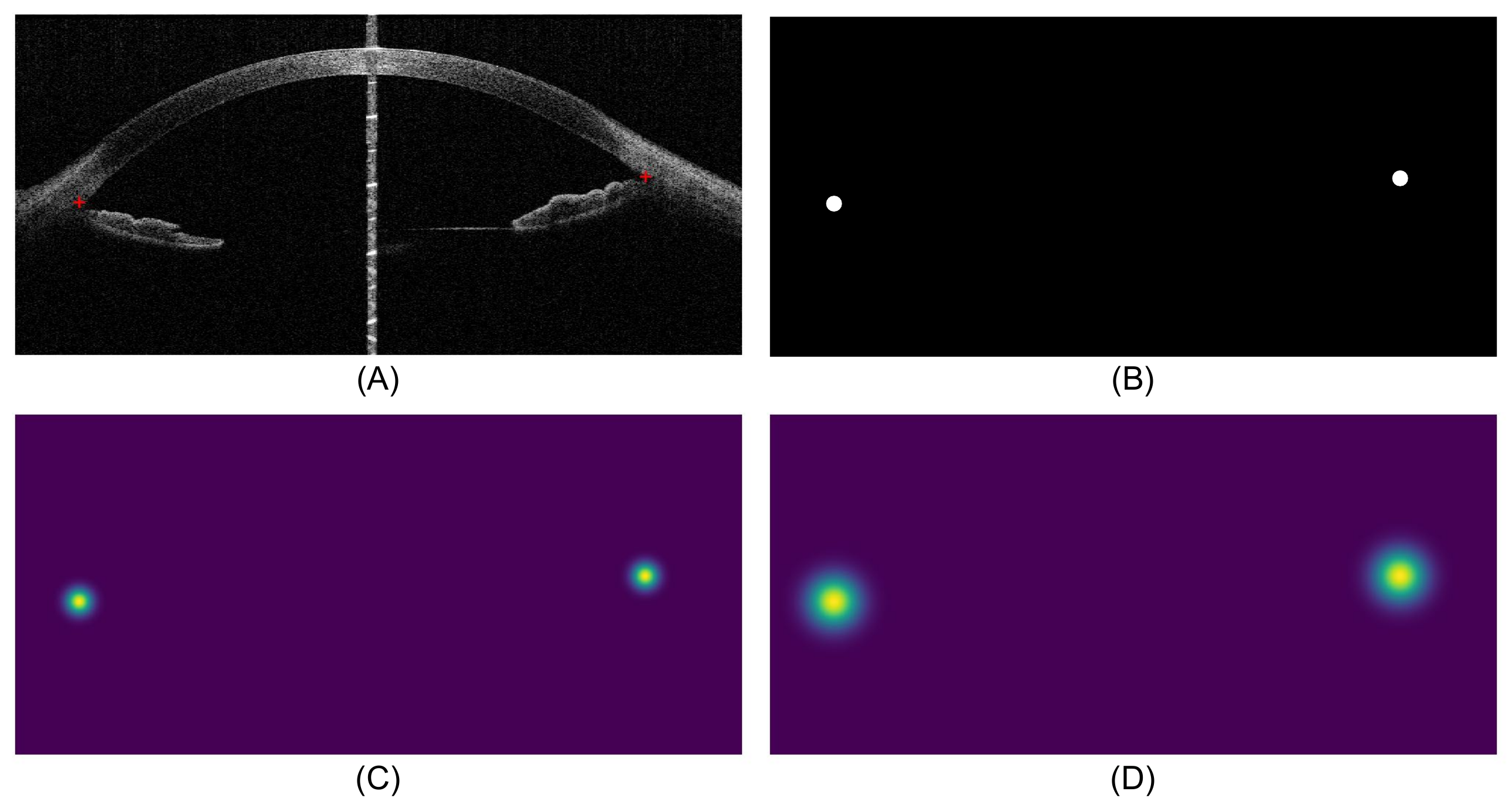}
	\caption{(A) Scleral spur localization aims to estimate the position of the scleral spur point from an AS-OCT image. (B) The binary mask based on the scleral spur region. (C, D) The heat maps generated based on scleral spur  with different radii.}
	\label{fig-SS_point}
\end{figure}

\paragraph{\textbf{Scleral spur localization task}} The aim is to estimate the position of the SS point from an AS-OCT image, as shown in Fig.~\ref{fig-SS_point} (A), which requires the algorithm to output the $(x, y)$ coordinates of the SS point in image coordinates.
Participants in the challenge proposed localization algorithms based on supervised learning, following one of three different approaches.
The first one was to directly predict the coordinates of the SS point, as a value regression problem. 
The second one was to extend the single pixel label to a small region, as shown in Fig.~\ref{fig-SS_point} (B). In this way, the SS localization task was transferred to a binary segmentation problem, where the segmented mask center was used as the SS position.  
The third approach was to generate a two-dimensional heat map based on the SS position, e.g., a Gaussian map, and then employ a regression method to estimate the SS point. With the heat map, the peak value was used as the SS position. Given the coordinates $(u_0, v_0)$ of the SS point, the heat map $G(u, v)$ could be calculated as:
\begin{equation}
G(u, v) = \exp \{ \dfrac{(u-u_0)^2 + (v-v_0)^2}{\delta ^2}\},
\end{equation}
where $\delta$ denotes the variance, a hyperparameter which controls the heat map radius. Fig.~\ref{fig-SS_point} (C, D) shows two generated Gaussian maps obtained with different values of $\delta$. 
Compared with the coordinate regression and binary segmentation approaches, the heat map solution reduces the complexity of the task, facilitating convergence during training. In addition, the method based on heat map regression can make use of a fully convolutional network for training and prediction.

\paragraph{\textbf{Angle closure classification task}} The aim is to predict the probability of a given AS-OCT image having a closed angle. Hence, the majority of participating teams built binary classification frameworks that would be suitable for the identification of angle closure.


In this section, we summarize these methods and analyze their corresponding results for the angle closure classification and scleral spur localization tasks. A brief summary of the methods are provided in Tables~\ref{tab_team_classification} and~\ref{tab_team_localization}, respectively.

\subsection{Cerostar Team}

\begin{figure}[!t]
	\centering
	\includegraphics[width = 1\linewidth]{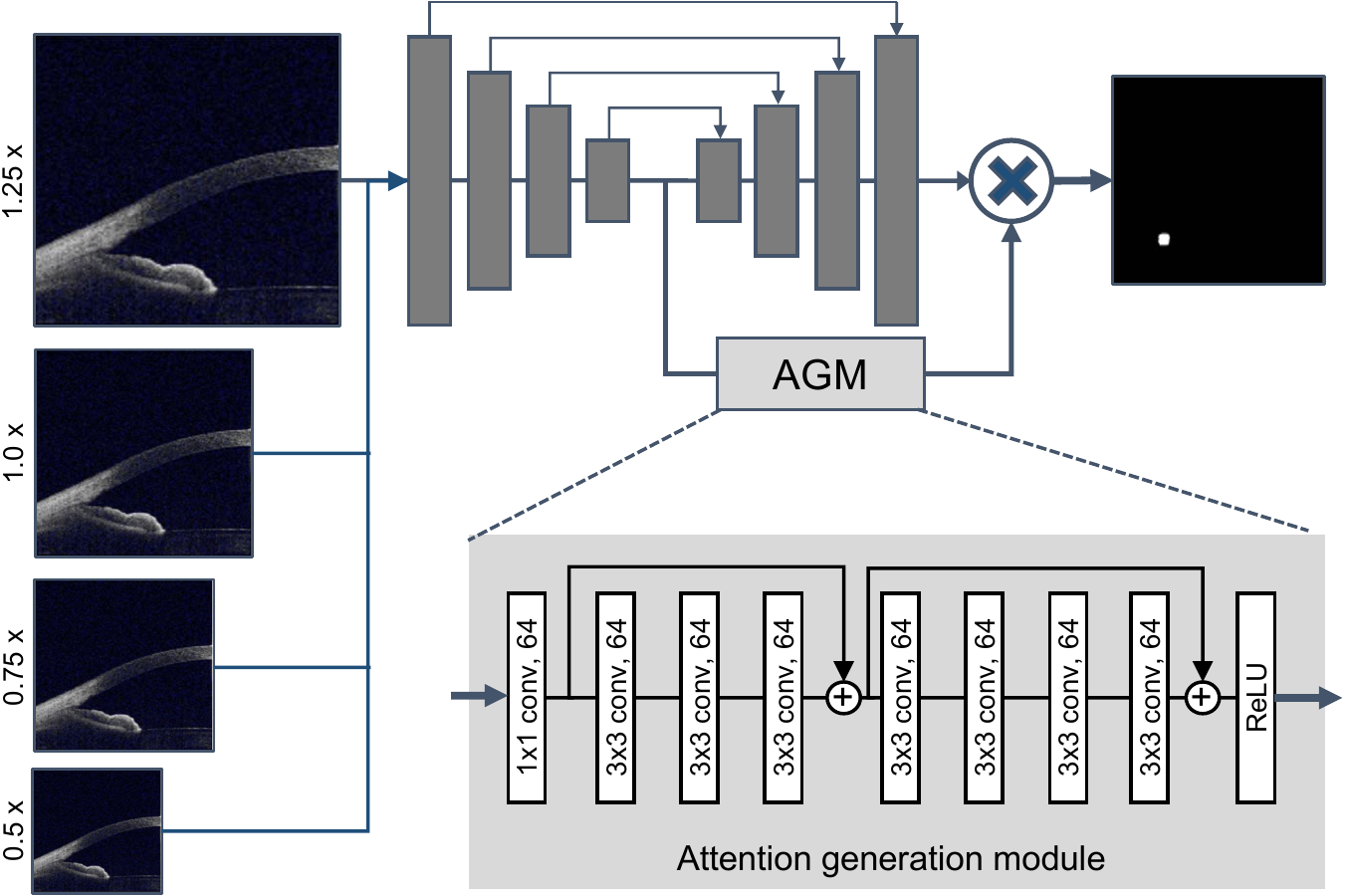}
	\caption{The framework of the Cerostar team for scleral spur localization, where an attention generation module was added into the backbone network.}
	\label{fig-Cerostar}
\end{figure}

\paragraph{\textbf{Scleral spur localization task}}
The Cerostar team utilized a multi-scale Res-UNet with an attention network as the backbone. As shown in Fig.~\ref{fig-Cerostar}, their proposed Res-UNet was based on a modified deep network ResNeXt34~\citep{ResNeXt} to extract semantic information from the input image. The Res-UNet contained a series of convolutional blocks composed of a convolutional layer, batch normalization layer, and ReLU activation. The last down-sampling layer in Res-UNet represented the semantic features of the image. The Cerostar team used four parallel Res-UNet with different sized images as inputs (i.e., 1.25 x, 1.0 x, 0.75 x, 0.5 x). Then, four semantic feature maps of the different sized Res-UNets were extracted and fed into the attention generation module, which contained eight CNN layers belonging to the first two blocks of ResNet34~\citep{ResNet2016} together with one ReLU layer. Finally, a weight matrix for each pixel was returned and used to combine the predictions of each Res-UNet and improve results.

\paragraph{\textbf{Angle closure classification task}}
The Cerostar team used a standard ResNet34 model~\citep{ResNet2016} pre-trained on ImageNet and fine-tuned using AGE training data to predict angle-closure glaucoma on the whole images.

\subsection{CUEye Team}

\begin{figure}[!t]
	\centering
	\includegraphics[width = 1\linewidth]{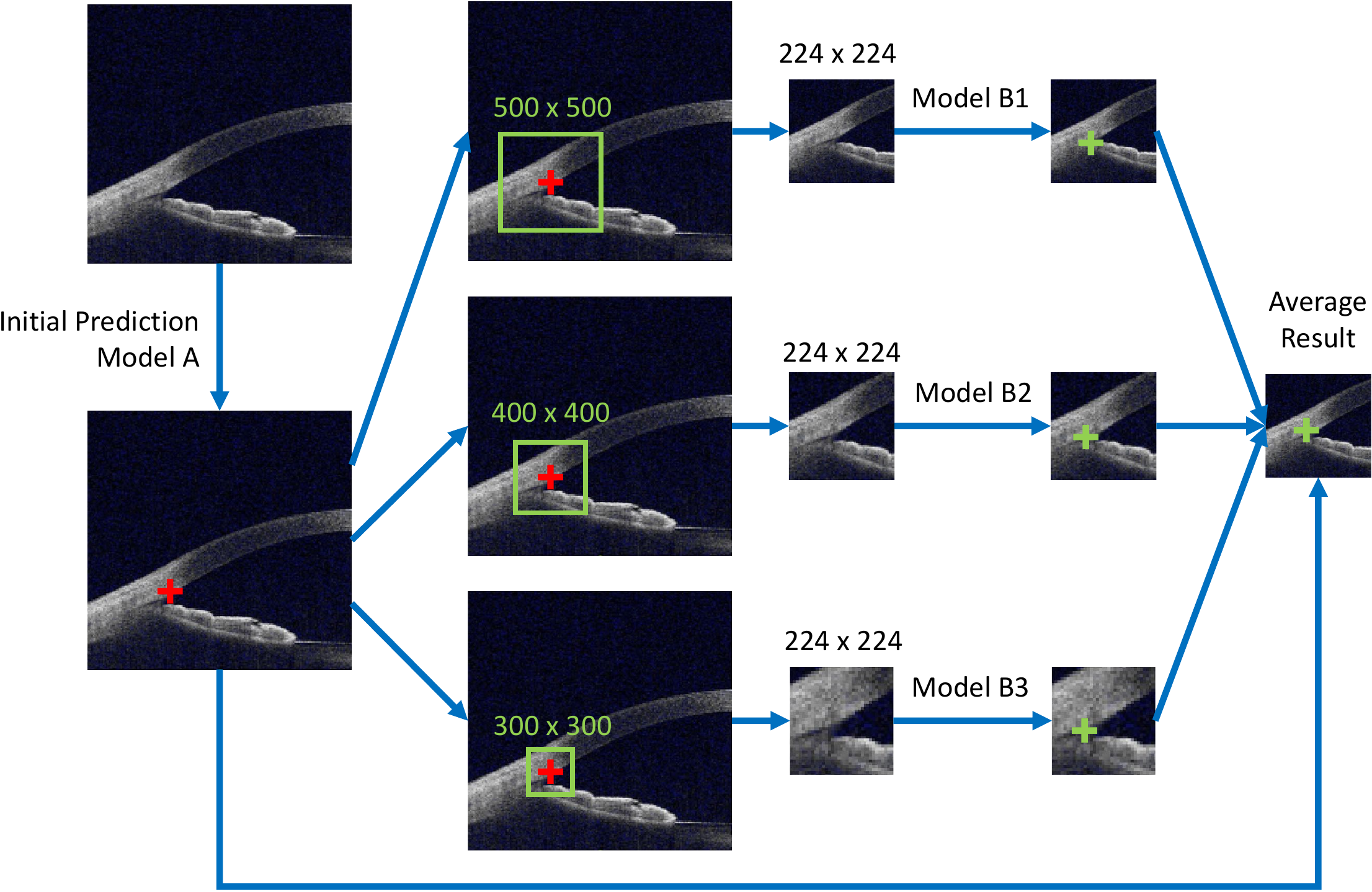}
	\caption{The network of the CUEye team for scleral spur localization, where a multi-scale pipeline was utilized to combine different ROIs. The SE-Net was used to predict the coordinates of the SS point.}
	\label{fig-CUEye}
\end{figure}

\paragraph{\textbf{Scleral spur localization task}}
The CUEye team employed a Zoom-in Squeeze-and-Excitation Network (SE-Net)~\citep{SENET}. Each AS-OCT image from the given dataset was split into two different parts according to the centerline, which simplified the problem to finding only one SS location in each given input. The AS-OCT images were captured with the specific position and direction~\citep{Fu2019_AJO}, so the CUEye team only used random shifting with a 0.2 scale, random rotation with 15 degrees, and random zooming with a 0.2 scale for image augmentation. 
Fig.~\ref{fig-CUEye} shows the framework of the CUEye team. An initial model $A$ was trained to make an initial prediction of the input based on SE-Net. Then, local regions of interest (ROIs) were randomly cropped from the original image into three regions of different sizes that covered approximately one-third to one-quarter of the original one around the initial prediction. Three parallel models $B1$, $B2$, and $B3$ with different input sizes were trained to make precise predictions based on a simple SE-Net module. Finally, the three parallel results were averaged together with the initial prediction to give the final results. 

\begin{figure*}[!t]
	\centering
	\includegraphics[width = 1\linewidth]{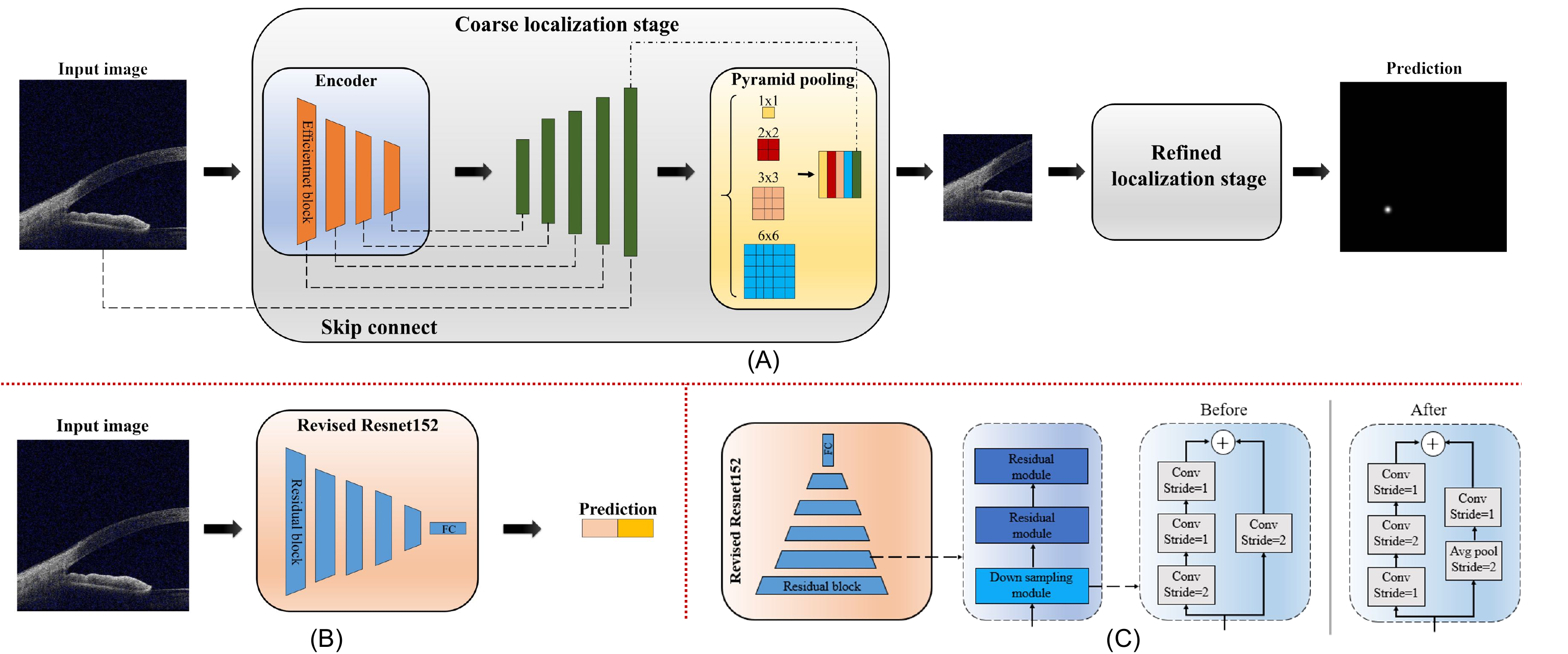}
	\caption{(A) The framework of the Dream Sun team for the scleral spur localization task, where the coarse and refined localization networks share the same model structure. (B) The framework of the Dream Sun team for angle closure classification. (C) The Revised ResNet structure.}
	\label{fig-DreamSun}
\end{figure*} 

\paragraph{\textbf{Angle closure classification task}}
The CUEye team employed a similar architecture as used for SS localization. An initial model $A$ was introduced to make the initial prediction of the scleral localization. Then, local ROIs were randomly cropped around the initial prediction into three regions of different sizes, and fed into three parallel SE-Net models to predict the classification results. Finally, their corresponding results were averaged to provide the final classification predictions.

\subsection{Dream Sun Team}

\paragraph{\textbf{Scleral spur localization task}}
The Dream Sun team introduced a coarse-to-fine strategy with progressive tuning, where the coarse and refined localization networks share the same model structure, as shown in Fig.~\ref{fig-DreamSun} (A). The point annotation of each SS was converted to a 2D Gaussian distribution map centered at the annotation position. EfficientNet~\citep{Tan2019} was chosen as the network encoder to learn and extract hierarchical features. After that, a skip connection module and a pyramid pooling module were utilized to capture the global and local semantic features from multiple dimensions and scales. 
Finally, the corresponding features were merged together to infer the final response regions. Considering the intensity and shape of SS regions, a combination of mean squared error (MSE) loss and Dice loss was used to reduce the error between prediction and ground truth. Specifically, the split images were resized to 499$\times$499 pixels for the coarse localization stage. Then, the candidate regions from the coarse stage were cropped to 360$\times$360 pixels for the precise localization stage. Considering efficiency and accuracy, the team selected EfficientNet-B2, B3, B5 and B6 to construct multiple models and then averaged these results to obtain the final prediction.

\paragraph{\textbf{Angle closure classification task}}
The Dream Sun team utilized ResNet152~\citep{ResNet2016} as the backbone architecture to perform accurate identification of angle closure. Two ResNet tweaks~\citep{8954382} were tailored to enhance the classification accuracy. 
To reduce the contextual information loss due to down-sampling in the first convolution with a stride of 2, they switched the strides of the first two convolutions, as shown in Fig.~\ref{fig-DreamSun} (C). 
Similarly, the residual connection mechanism in the down-sampling module also ignored $3/4$ of the input feature maps. Empirically, a 3$\times$3 average pooling layer with a stride of 2 was inserted before the convolutional layer.  To tackle the class imbalance problem between angle closure and non-closure samples, a hybrid loss combining the Focal loss~\citep{Lin_2017_ICCV} and F-beta loss~\citep{pmlr-v54-eban17a} was adopted. Each OCT image was symmetrically split into two sub-images (left and right) and resized to 256$\times$256 pixels for identifying the angle status. In addition, the training dataset was further augmented with random re-scaling, flipping and rotation. The Adam optimizer and cosine learning rate decay strategy were adopted to update the network weights. The final result was decided by the majority vote of three models established with different training iterations.

\subsection{EFFUNET Team}

\begin{figure}[!t]
	\centering
	\includegraphics[width = 1\linewidth]{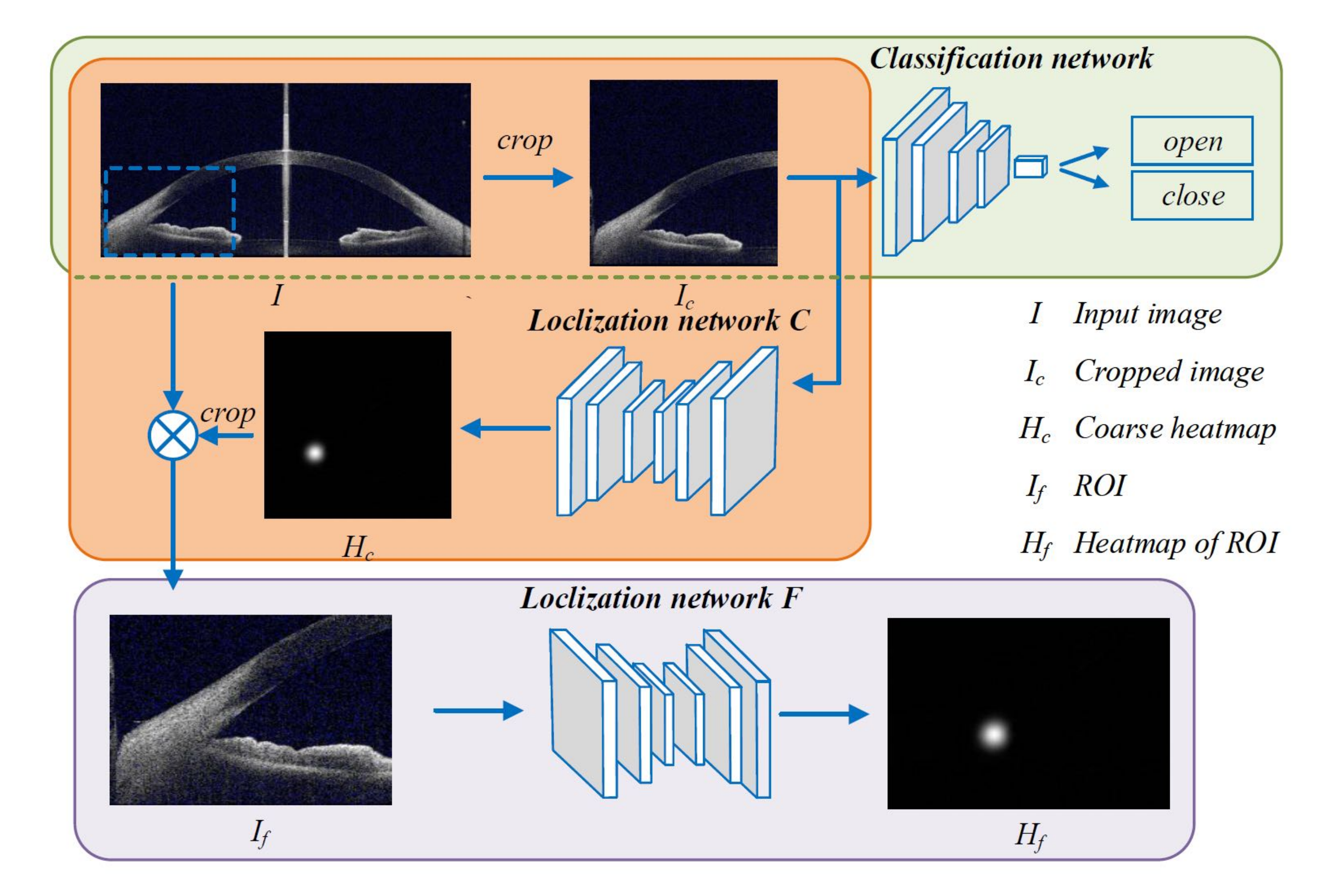}
	\caption{The framework of the EFFUNET team for the two tasks.}
	\label{fig-EFFUNET_frame}
\end{figure}

\paragraph{\textbf{Scleral spur localization task}}
The EFFUNET team proposed a coarse-to-fine localization framework~\citep{Wang2019g}, which consisted of two networks with the same architecture, as shown in Fig.~\ref{fig-EFFUNET_frame}. These models were trained using heat map regression, where each network was first regressed against the ground truth heat map at a pixel level and then the predicted heat maps were used to infer landmark locations. The coarse network was trained to delineate the coarse localization heat maps and to generate the ROI of the key-points, while the fine network was trained for accurate localization using the cropped ROI from the whole OCT slices. A U-Net structure~\citep{Falk2019} was adopted as the backbone architecture for each of these two components. Their encoders were based on EfficientNet-B5~\citep{Tan2019}, as its scaling method allows networks to focus on more relevant regions with object details. The MSE loss was used to supervise the regression of the heat maps.

\paragraph{\textbf{Angle closure classification task}} 
The EFFUNET team employed EfficientNet~\citep{Tan2019} as their backbone. Using a simple and highly effective compound scaling method, EfficientNet achieved state-of-the-art accuracy on the ImageNet dataset. As the resolution of the original images is 2030$\times$998, each image was cropped into left and right images (998$\times$998) along the corresponding vertical center line. The cropped images were resized to 384$\times$384 for the classification network. The final classification result was assigned by averaging the outputs of EfficientNet-b3 and EfficientNet-b5.

\subsection{iMed Team}

\begin{figure*}[!t]
	\centering
	\includegraphics[width = 1\linewidth]{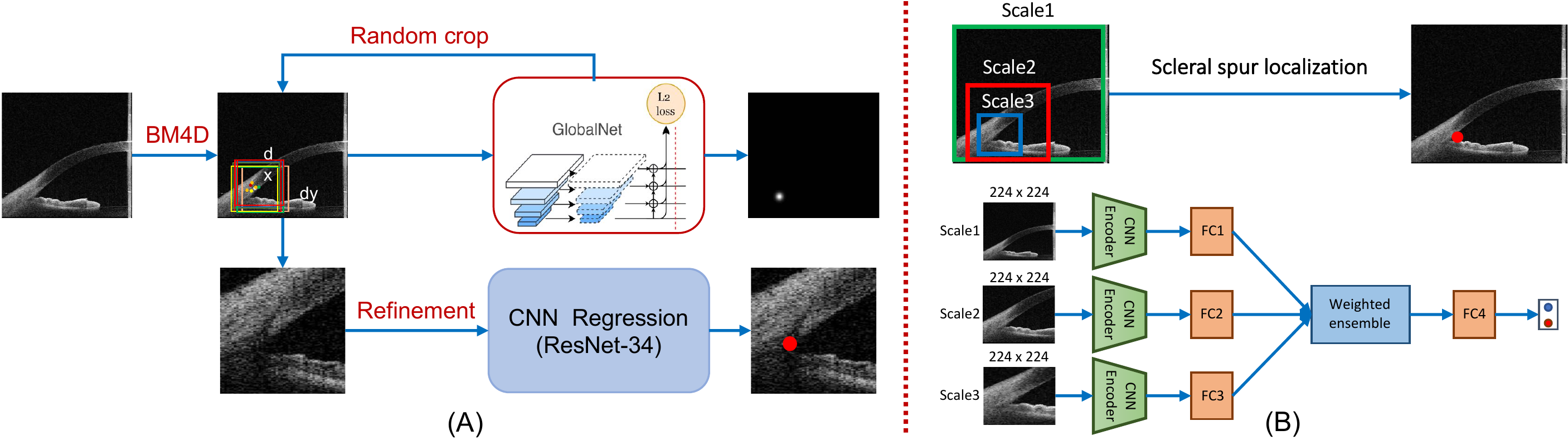}
	\caption{(A) The framework of the iMed team for scleral spur localization, where GlobalNet was adopted as the backbone. (B) The framework of the iMed team for angle closure classification, where a multi-scale network was employed to fuse multi-scale ROIs.}
	\label{fig-imed_model}
\end{figure*}

\paragraph{\textbf{Scleral spur localization task}}
The iMed team also employed a coarse-to-fine framework, as shown in Fig.~\ref{fig-imed_model} (A). In order to improve the performance, an image denoising method, BM4D~\citep{maggioni2012nonlocal}, was first employed to suppress the background noise. GlobalNet, with a cascaded pyramid network~\citep{chen2018cascaded}, was used for coarse position localization. Next, a random cropping processing was applied, in which the team randomly chose a point in a square region with sides of $30$ pixels and then, keeping its relative position, cut out a 224$\times$224 patch from the image. These ROIs were then input to the CNN regression network to obtain the SS localization results. The pre-trained ResNet-34~\citep{ResNet2016} was employed as the backbone architecture. First, ACA regions of 224$\times$224 were fed into the network to extract a finer feature representation. Then, since the ground truth of the localization was normalized between 0 and 1, a Sigmoid activation function was appended to the fully connected layer to normalize the coordinate values of the output. The MSE loss was chosen to supervise the training process.

\paragraph{\textbf{Angle closure classification task}}
The iMed team used a multi-scale network with ResNet-50~\citep{ResNet2016} as the backbone, with three-scale inputs in addition to the original scale, and cropped ROIs of sizes 448$\times$448 and 224$\times$224, as shown in Fig.~\ref{fig-imed_model} (B).  In clinical practice, the ACA region is the most important sign for diagnosis of glaucoma type. The global image ($Scale1$) with complete AS-OCT structure could provide more global information. Meanwhile, the local images, $Scale2$ and $Scale3$, preserve  local details  with  higher resolutions and were thus used to learn a fine representation. 
Three regions with different sizes were scaled to 224$\times$224 and used to learn different feature representations output from the last convolutional layers in  ResNet-50~\citep{ResNet2016}.  The 7$\times$7 feature maps from the parallel network modules were fed into a global max pooling layer.
A set of different descriptors from each stream was obtained, where each 2$\times$1 descriptor was generated by the fully connected layers in the classification network. To obtain the best prediction result, the descriptors were concatenated to create a new descriptor with size 2$\times$3. 
A convolution operation with 32 kernels of size 1$\times$3 was applied to the new descriptor, and then the results were fed to the fully connected layer for final classification. The 1$\times$3 kernels weighted the predictions of the three models and output them to the next layer. This feature ensemble strategy enabled the models to automatically learn the importance of different basic predictions. Finally an objective function following a multi-scale loss $L_m$ was used, as given by:
\begin{equation}
	L_m = \sum_{s=1}^{3}\left \{ L_{cls}\left ( y^{(s)}, y^{*} \right ) \right \}+L_{cls}(y_{f}, y^{*}),
\end{equation}
where $s$ denotes each scale, and $y^{(s)}$ and  $y^{*}$ denote the predicted label vector from a specific scale and the ground truth label vector, respectively. $y_{f}$  denotes the final predicted vector from the three-scale feature ensemble. $L_{cls}$ represents the classification loss, e.g., CE loss, which predominantly optimizes the parameters from the convolutional and classification layers.

\subsection{MIPAV Team}

\begin{figure}[!t]
	\centering
	\includegraphics[width = 1\linewidth]{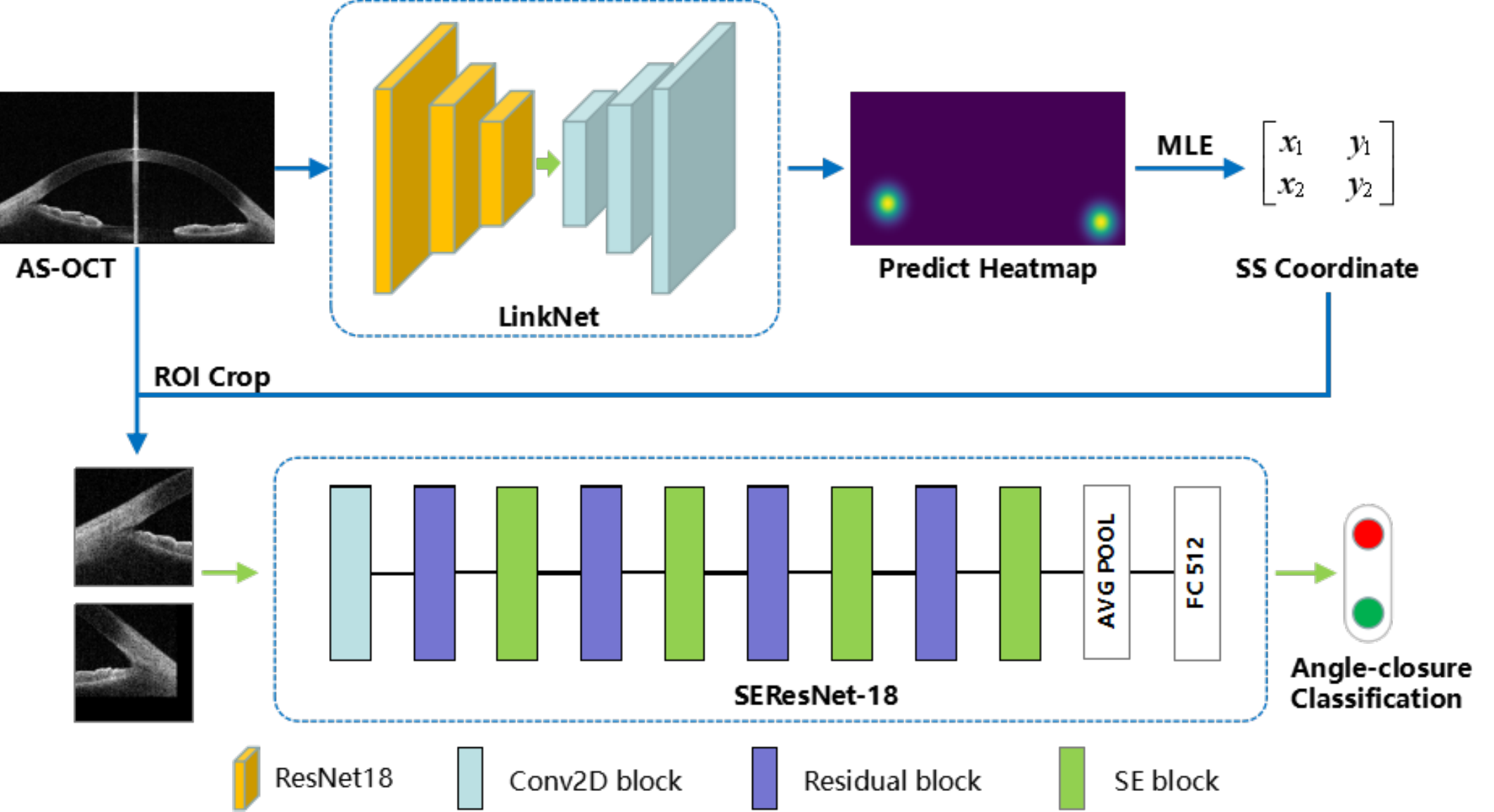}
	\caption{Schematic representation of the MIPAV team's method for scleral spur localization and angle closure classification. LinkNet was first trained to generate the heat map to localize the scleral spur. The information contained in the heat map was extracted through a method developed from maximum likelihood estimation (MLE) to obtain the scleral spur coordinates. Then, the ACA patches were cropped with the scleral spur centered, and fed into the classification model (i.e., SE-ResNet) to classify angle closure.}
	\label{fig-MIPAV_frame}
\end{figure}

\paragraph{\textbf{Scleral spur localization task}}
The MIPAV team employed LinkNet~\citep{LinkNet}, a typical light U-shaped architecture, to learn the transformation from an AS-OCT image to a probability map, as shown in Fig.~\ref{fig-MIPAV_frame}. The encoder part was based on a pre-trained model of ResNet18~\citep{ResNet2016}, which retained the first four extraction blocks without the average pooling or fully connected layers. 
Compared with U-Net~\citep{Falk2019}, LinkNet uses an addition operation rather than a concatenation for the skip connection, which can reduce the computational cost and accelerate the training process. To learn the pixel-wise regression network, the MSE loss was utilized to calculate the difference between the ground truth and predictions. Random data augmentation was applied before training, including adjusting brightness, contrast and sharpness. All enhancement factors followed a log-normal distribution. 
Moreover, the MIPAV team considered the pixel value of the generated heat map as an ideal 2D Gaussian probability density. 
A method based on maximum likelihood estimation (MLE) theory was developed to obtain the coordinates from this output heat map, which is defined as follows:
\begin{equation}
u_c = \dfrac{\sum_{i \in C}u_i p_i}{\sum_{i \in C} p_i} , \;\; v_c = \dfrac{\sum_{i \in C}v_i p_i}{\sum_{i \in C} p_i} ,
\end{equation}
where $C$ indicates the set containing pixels whose values are higher than  half of the maximum in the heat map. $(u_i, v_i)$ are the corresponding coordinates, and $p_i$  is the value of the pixel in the set. Basing the results on the weighted average operation produces less error than finding the peak directly. 

\paragraph{\textbf{Angle closure classification task}}
The MIPAV team used a modified SE-ResNet18~\citep{SENET} as the backbone. The SS coordinates were utilized to localize the ACA region. A 128$\times$128 patch with the SS centered was cropped as the input of the classification network. To reduce the localization error, noise following a Gaussian distribution was added to the real coordinates when cropping patches. The operation also served as a form of data augmentation to make the model more generalizable. 
A SE-ResNet18~\citep{SENET} was modified as the backbone. Experiments in~\citep{SENET} showed that integrating the SE block in different positions of the residual blocks achieves similar results. As shown in Fig.~\ref{fig-MIPAV_frame}, the SE block was inserted between each residual layer of a pre-trained ResNet18~\citep{ResNet2016}, without destroying the original residual architecture. Data imbalance was another difficulty for achieving accurate classification. To overcome this problem, the Focal loss~\citep{Lin_2017_ICCV} was employed as the cost function during training, given by:  
\begin{equation}
L_{focal} = - \alpha (1 - \hat{y})^{\gamma}  \log (\hat{y}) - (1 - y^{*}) \hat{y}^{\gamma}  \log (1 - \hat{y}),
\end{equation}
where $\hat{y}$ and $y^{*}$ denote the predicted label and ground truth, respectively. $\alpha$ and $\gamma$  are weighted parameters ($\alpha = 6$ and $\gamma = 2$). After analyzing the labels, the MIPAV team found that the closure status of both the left and right angle in the same AS-OCT image were directly correlated. As such, a voting mechanism was introduced in the final test, which efficiently increased the accuracy of classification. 

\subsection{RedScarf Team}

\begin{figure*}[!t]
	\centering
	\includegraphics[width = 1\linewidth]{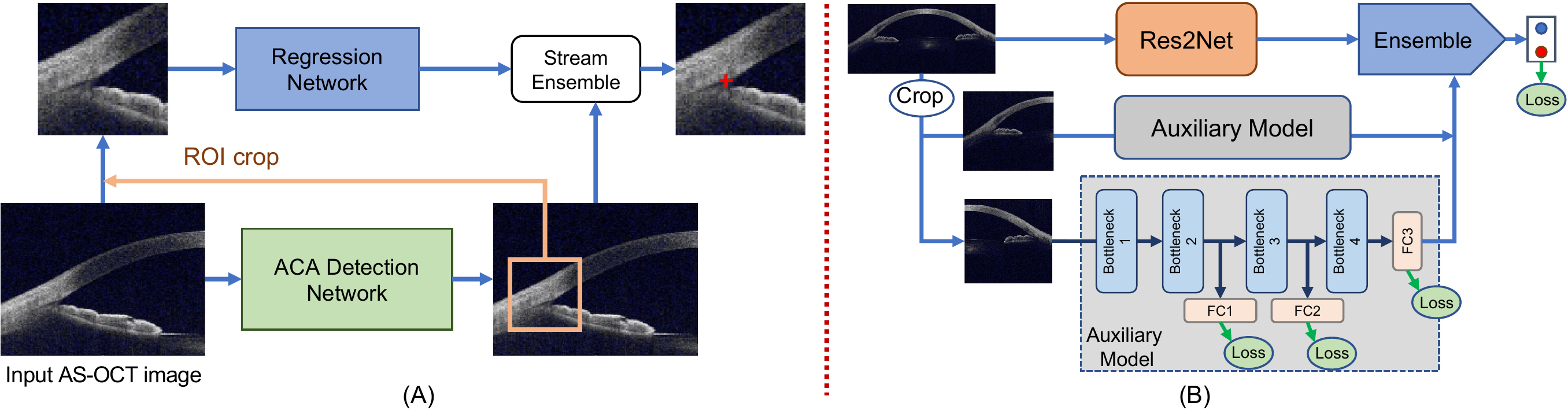}
	\caption{(A) The framework of the Redscarf team for scleral spur localization. The ACA detection network  was used to predict the initial ROI, while a regression network was employed to produce accurate scleral spur coordinates. (B) The framework of the Redscarf team for angle closure classification, where a multi-scale network was employed to fuse three different scales of ROIs. }
	\label{fig-Redscarf_model}
\end{figure*}

\paragraph{\textbf{Scleral spur localization task}}
The Redscarf team proposed a two-stream framework combining ROI detection and heat map regression, as shown in Fig.~\ref{fig-Redscarf_model} (A). The ROI detection was sensitive to the ACA structure, but the localization accuracy was not adequate. In contrast, the heat map regression had a high localization accuracy, but it was easily affected by noise, misdetecting noisy points located far away from the true scleral spur pixel. 
As such, the team first exploited YOLO-V3~\citep{redmon2018yolov3} as the detection network to identify the ROI region. Then, AG-Net~\citep{AGNet_2019} was used as a regression network to produce accurate coordinate values. Compared with U-Net~\citep{Falk2019}, AG-Net replaces the concatenation with an attention guided filter to enhance the skip connection, which can reduce the influence of noise and accelerate the testing process. 
The detection network may identify several ACA candidates, so the team proposed to average them to produce the final prediction of the true ACA structure center. Centered on the ACA structure center, a 64$\times$64 patch was cropped as the ROI region. To train the heat map regression network, the MSE loss was utilized to calculate the difference between ground truth and prediction. In order to improve the generalization capabilities of the model, random rotations of $[-10, 10]$ degrees were applied before training. To convert the output heat map to the final scleral spur localization result, they averaged the position of values greater than a threshold in the map.

\paragraph{\textbf{Angle closure classification task}}
The RedScarf team employed a three-branch network based on Res2Net~\citep{gao2019res2net} as the backbone, as shown in Fig.~\ref{fig-Redscarf_model} (B). Compared with ResNet~\citep{ResNet2016}, Res2Net further constructs hierarchical residual-like connections within a single residual block, which enables multi-scale features to be better captured. First, the image was cropped into two sections, which were fed to an auxiliary model to obtain an early prediction. The auxiliary model contained four bottlenecks with three extra auxiliary losses. Since the auxiliary losses were not equally important, they imposed different confidences over them (0.2, 0.3, 0.5). Training models with the auxiliary loss has three advantages: 1) It encourages features in the lower level to be more discriminative. 2) It alleviates the gradient vanishing problem in the lower level of the model. 3) It provides additional regularization. In order to model the relationship between two angles, the classification of the whole image was also treated as a multi-label task using Res2Net. Finally, the classification result was obtained by combining the three outputs above.

\subsection{VistaLab Team}

\begin{figure*}[!t]
	\centering
	\includegraphics[width = 1\linewidth]{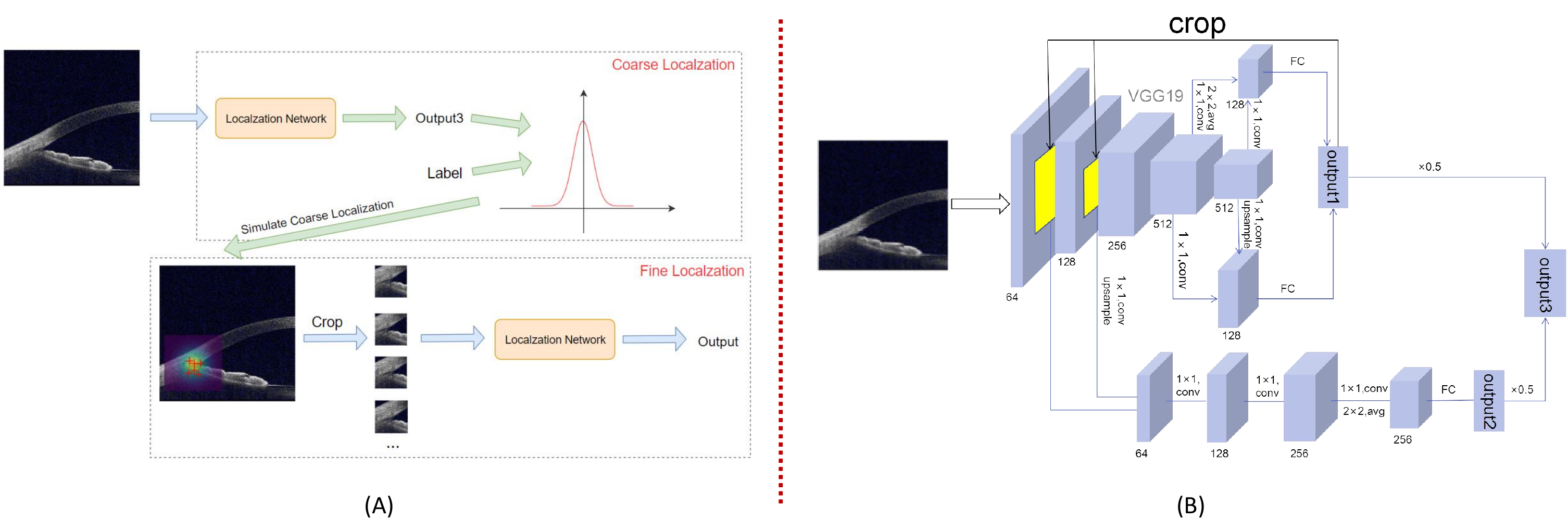}
	\caption{(A) The framework of the VistaLab team for scleral spur localization, where a coarse-to-fine framework was used to detect the scleral spur points. (B) The localization sub-network of the VistaLab team used in this framework. }
	\label{fig-VistaLab_model}
\end{figure*}

\paragraph{\textbf{Scleral spur localization task}}
The VistaLab team employed a coarse-to-fine framework, as shown in Fig.~\ref{fig-VistaLab_model} (A). The localization network was based on a pre-trained VGG19~\citep{VGG}. In deep neural networks, large-scale feature maps often contain more details and contour information, which may be helpful for position detection. 
Therefore, they extracted more feature maps from the fourth block of VGG19, which were twice as large as those from the fifth block. The outputs of both blocks were fused to obtain more effective features. Based on a 1$\times$1 convolutional layer, the dimensions of the fused features were reduced for  more compactness with less parameters. Then the feature maps of the fourth block were down-sampled and combined with the feature maps of the fifth block. At the same time, they also up-sampled the feature maps of the fifth block to combine them with the feature maps of the fourth block. Based on this, the feature maps of the fourth block  contained more semantic information, while the fused feature maps of the fifth block  contained more details and contour information. Finally, the feature maps of these two blocks were passed through the fully connected layer in sequence to produce two sets of coordinates. The coarsely positioned coordinates (Output1) were obtained by averaging these two sets of coordinates, as shown in Fig.~\ref{fig-VistaLab_model} (B). 
Output1 was then used to crop the feature maps of the first block and  the second block of VGG19, with ROIs of 16$\times$16 and 8$\times$8, respectively. The cropped feature maps were then further fused by a sub-network, which consisted of three convolutional blocks and a fully connected layer, generating another position result (Output2). Finally, Output1 and Output2 were averaged to get the final coarse position result (Output3).  Based on the  coarse positioning network, the random   224$\times$224 ROIs with a Gaussian distribution were cropped. These ROIs and labels were used to  train the fine positioning network, which had the same architecture as the coarse positioning network. The ED loss was chosen to train the coarse-to-fine framework.

\paragraph{\textbf{Angle closure classification task}}
The VistaLab team utilized ResNet18~\citep{ResNet2016} as the backbone, and changed the output of the last linear layer to 2.  The model was trained using the Adam optimizer and CE loss function. Considering that the classified data is not balanced, four-fold training was used to objectively evaluate the model. The dataset was proportionally divided into four parts, one of which was used as test data, while the remaining data was used for training. According to different divisions, four different models were trained respectively. Then, the models that performed best on both the training set and test set, in terms of highest accuracy and lowest loss, were selected. Finally, the four results were averaged to get the final classification result.

\section{Results and Discussion}

In this section, we report the evaluation metrics of the eight teams on  both the online (validation) and onsite (testing) datasets for the two proposed tasks. The leaderboards of the overall challenge can be accessed on the AGE challenge website at \curl{https://age.grand-challenge.org}.

\subsection{Task 1: Scleral Spur Localization}

\begin{figure}[!t]
	\centering
	\includegraphics[width = 1\linewidth]{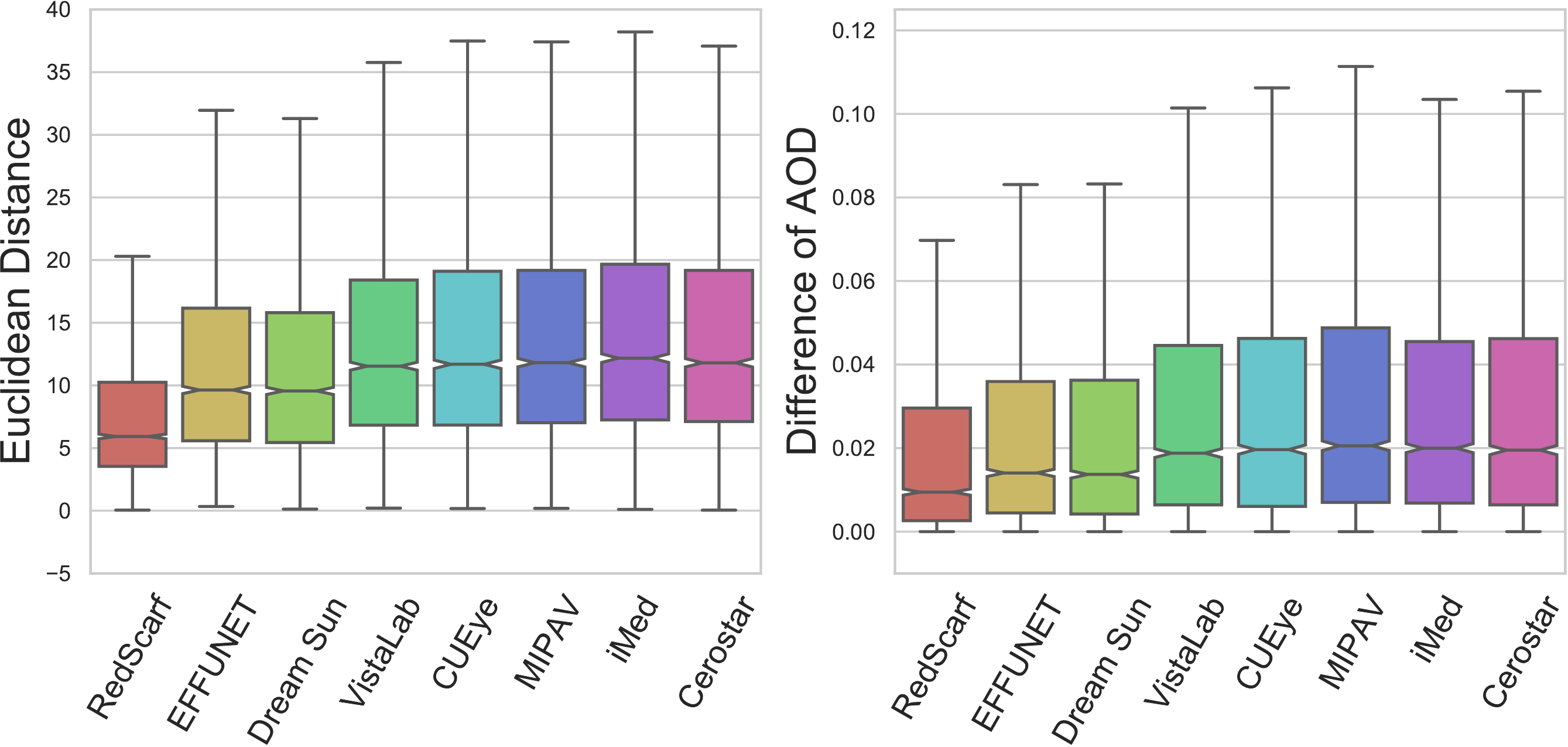}
	\caption{Box-plots illustrating the scleral spur localization performances on the onsite dataset. Left: Euclidean Distance. Right: Differences of Angle Opening Distance (AOD). }
	\label{fig-boxplot}
\end{figure}

\begin{figure*}[!t]
	\centering
	\includegraphics[width = 1\linewidth]{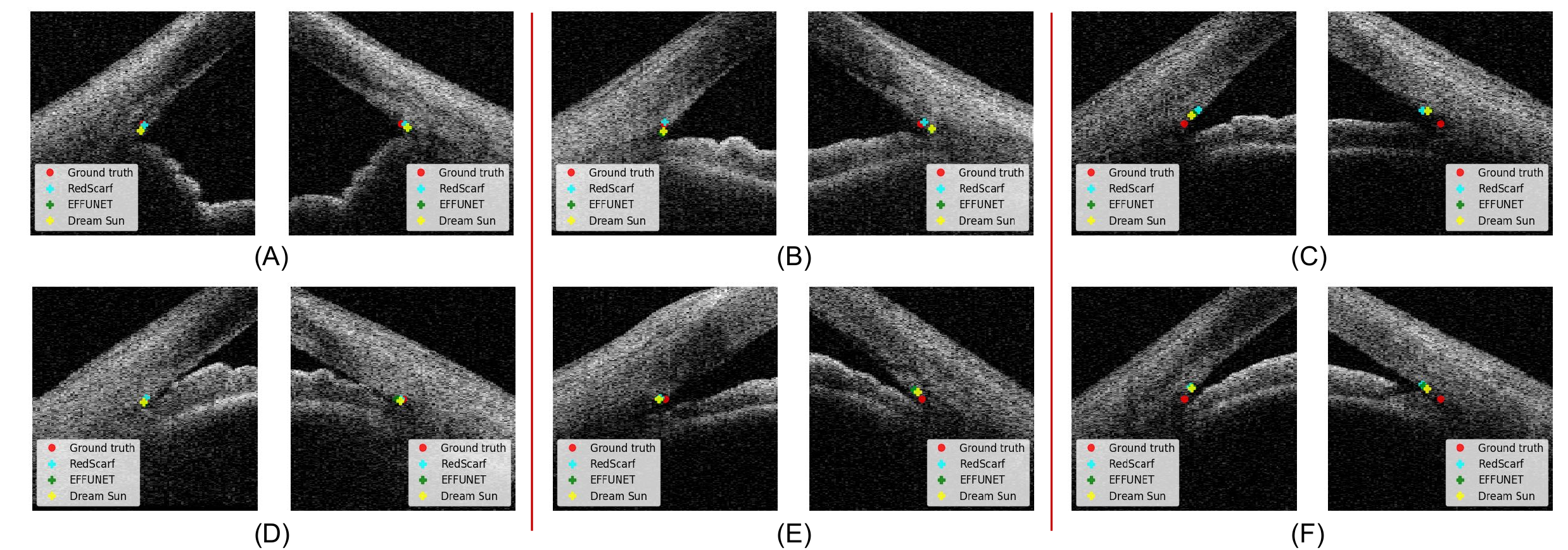}
	\caption{Zoomed-in scleral spur localization results of ground truth and top-three teams (i.e., RedScarf, EFFUNET and Dream Sun). Top raw images are open angle cases, while bottom raw images are angle-closure cases. }
	\label{fig-result}
\end{figure*}

Results and box-plots summarizing the distribution of the performance metrics obtained by each of the participating teams for the scleral spur localization are presented in Table~\ref{tab_result_localization} and Fig.~\ref{fig-boxplot}, respectively. The RedScarf team achieved the best performance on the onsite dataset with $ED$ of 9.395 and $\Delta_{AOD}$ of 0.02772. The EFFUNET team ($ED$ of 12.653 and $\Delta_{AOD}$ of 0.03546) and Dream Sun team ($ED$ of 12.468 and $\Delta_{AOD}$ of 0.03654) achieved the second and third best performances, respectively, on the onsite dataset. 
Fig.~\ref{fig-result} shows several qualitative examples of the  scleral spur localization of the top-three ranked methods (i.e., RedScarf, EFFUNET and Dream Sun) together with the ground truth, where (A-C) are open angle images, and (D-F) are angle-closure images. The general behaviors of all methods were fairly stable relative to each other, in most cases. Fig.~\ref{fig-result} (C, F) illustrates some challenging low-quality images, where the poor illumination and low-contrast (e.g., Fig.~\ref{fig-result} (C)) often made it difficult to determine the SS point.


\begin{table*}[!t]
	\caption{Results of the scleral spur localization task.}
	\small 
	\centering
	\begin{tabular}{|l|c|c|c||c|c|c||c|c|}
		\hline
		Team      & \multicolumn{3}{c||}{ Online (validation) data  } & \multicolumn{3}{c||}{ Onsite (testing) data  } & \multicolumn{2}{c|}{ Final Ranking } \\ \cline{2-9}
		          & $ED$  & $\Delta_{AOD}$ &           Rank           & $ED$  & $\Delta_{AOD}$ &         Rank          & $S_{final}$ &          Rank          \\ \hline \hline
		EFFUNET   & 12.95 &     0.0456     &            4             & 12.65 &     0.0355     &           2           &     2.4     &           1            \\ \hline
		RedScarf  & 16.55 &     0.1200     &            8             & 9.40  &     0.0277     &           1           &     2.4     &           1            \\ \hline
		Dream Sun & 12.90 &     0.0424     &            1             & 12.47 &     0.0365     &           3           &     2.6     &           3            \\ \hline
		VistaLab  & 15.18 &     0.0470     &            6             & 14.00 &     0.0430     &           4           &     4.4     &           4            \\ \hline
		CUEye     & 13.43 &     0.0450     &            3             & 14.39 &     0.0430     &           5           &     4.6     &           5            \\ \hline
		MIPAV     & 13.76 &     0.0390     &            2             & 14.35 &     0.0469     &           6           &     5.2     &           6            \\ \hline
		iMed      & 16.32 &     0.0547     &            7             & 14.87 &     0.0483     &           7           &     7.0     &           7            \\ \hline
		Cerostar  & 13.53 &     0.0472     &            5             & 14.41 &     0.0486     &           8           &     7.4     &           8            \\ \hline
	\end{tabular} 
	\label{tab_result_localization}%
\end{table*}

\subsection{Task 2: Angle Closure Classification}

\begin{table*}[!t]
	\caption{Results of the angle closure classification task.} 
	\small 
	\centering
	\begin{tabular}{|l||c|c|c|c||c|c|c|c||c|c|}
		\hline
		Team      & \multicolumn{4}{c||}{ Online (validation) data  } & \multicolumn{4}{c||}{ Onsite (testing) data  } & \multicolumn{2}{c|}{ Final Ranking } \\ \cline{2-11}
		          &   AUC   & Sensitivity & Specificity &    Rank     &   AUC   & Sensitivity & Specificity &   Rank   & $S_{final}$ &         Rank         \\ \hline \hline
		EFFUNET   & 1.00000 &   1.00000   &   1.00000   &      1      & 1.00000 &   1.00000   &   1.00000   &    1     &  1.0  &          1           \\ \hline
		RedScarf  & 0.99976 &   0.99375   &   0.99531   &      8      & 1.00000 &   1.00000   &   1.00000   &    1     &  2.4  &          2           \\ \hline
		VistaLab  & 1.00000 &   0.99688   &   1.00000   &      4      & 0.99998 &   1.00000   &   0.99375   &    3     &  3.2  &          3           \\ \hline
		Dream Sun & 1.00000 &   1.00000   &   1.00000   &      1      & 0.99992 &   1.00000   &   0.98750   &    4     &  3.4  &          4           \\ \hline
		MIPAV     & 1.00000 &   1.00000   &   1.00000   &      1      & 0.99992 &   0.99688   &   0.99844   &    6     &  5.0  &          5           \\ \hline
		iMed      & 0.99983 &   0.98750   &   0.99844   &      7      & 0.99959 &   1.00000   &   0.99375   &    5     &  5.4  &          6           \\ \hline
		Cerostar  & 0.99999 &   0.99959   &   1.00000   &      5      & 0.99491 &   1.00000   &   0.97422   &    7     &  6.6  &          7           \\ \hline
		CUEye     & 0.99297 &   1.00000   &   0.98594   &      6      & 0.98203 &   1.00000   &   0.96406   &    8     &  7.6  &          8           \\ \hline
	\end{tabular} 
	\label{tab_result_classification}%
\end{table*}

The participating methods for the angle closure classification task are reported in Table~\ref{tab_result_classification}. As can be observed, the RedScarf and EFFUNET teams obtained  perfect scores on the onsite dataset. Further, almost all the methods achieved a Sensitivity of 100\%, while the major differences in performance between the methods is seen in the Specificity scores, ranging from $96.4\%$ to $100\%$. 
There are several possible reasons for this high-performance: 
1) The angle closure cases in the AGE challenge were at moderate or advanced stage, with an obvious closed anterior chamber angle making them easy to discriminate from open angle cases. 
2) In contrast to the  quantitative clinical measurements (e.g. anterior chamber area, ACW, AOD, and angle recess area), the visual representations extracted by deep networks can present more information beyond what clinicians recognize as relevant. This point was also observed in other angle closure studies~\citep{Fu2019_AJO,XU2019_AJO,Fu2019_ASOCT_TC}.

\subsection{Discussion}

From the AGE challenge results, the top-performing approach (RedScarf) had an average ED of 10 pixel (10$\mu$m) in scleral spur localization, while in the task of angle closure classification, all the algorithms achieved satisfactory performances,with the top-two (EFFUNET and RedScarf) obtaining accuracy rates of 100\% on the onsite dataset. In this section, we provide more analysis and discussion comparing the different solutions. 

\subsubsection{Scleral Spur Localization Task}
 
The original resolution of AS-OCT images was 2130$\times$998,  which is too large for training deep models directly due to limitations in GPU memory. As such, most teams chose a coarse-to-fine strategy. For example, the solutions of the top-three teams were all based on an ROI cropped flowchart. In fact, six out of the eight teams employed this strategy (see Table~\ref{tab_team_localization}) to ensure a more precise localization. This was perhaps motivated by the fact that the scleral spur labels were provided as single pixels. Therefore, identifying a first approximation of the area and predicting the final value in a second iteration allows more detailed features of the ACA structure to be preserved and prevents information loss caused, for example, by down-sampling. The standard U-Net was utilized by most methods for identifying the initial ROI, which could provide a satisfactory result.

As mentioned in Section~\ref{sec-method}, there are three main solutions for SS localization from ROI and global images, \eg, value regression,  binary mask segmentation, and heat map prediction. Value regression directly predicts the coordinates of the SS point by using a deep regression network (\eg, SE-Net~\citep{SENET}, ResNet~\citep{ResNet2016}, or VGG~\citep{VGG}). However,  CNN regression networks have more parameters than the segmentation networks, due to the fully connected layers, and thus require more training data to avoid overfitting. 
Moreover, for a high-resolution image (\ie, 2130$\times$998), it is challenging to predict  accurate coordinates within a small pixel-level range. 
By contrast, the binary mask segmentation and heat map prediction are based on segmentation networks (\eg, U-Net~\citep{Falk2019} or AG-Net~\citep{AGNet_2019}), which can be optimized well with limited data.
Moreover, heat maps can extend the scleral spur position from a single pixel to a small area that can easily be approximated. From the online challenge evaluation results, it can be deducted that modeling the SS localization task as a heat map prediction problem is appropriate, with the top-three teams being based on this.
 	 
In terms of network architecture, the Cerostar and CUEye teams utilized a multi-scale ensemble framework to integrate the different input ROIs. The Dream Sun team employed a multi-model fusion strategy to combine  the results of EfficientNet B2, B3, B5, and B6. Other teams based their methods on a single model to predict the SS localization. From the challenge evaluation results, the ensemble strategy does not gain a significant improvement over the single models. One possible reason is that, as a one-pixel position prediction, semantic information in a larger view does not provide more representations than that in a small ROI for SS point localization.  
Overall, the single segmentation models based on ROIs can achieve satisfactory results for the scleral spur localization task.

\subsubsection{Angle Closure Classification Task}
 
Similar to the scleral spur localization task, coarse-to-fine strategies were also widely used for angle closure classification (five out of eight teams in Table~\ref{tab_team_classification}). The general flowchart was first to identify the SS point and then crop a smaller ROI, which was fed to a classification network to predict the angle closure. One major reason for doing this is that the main representations used to describe features of the anterior chamber angle fall into the ACA region, which is consistent with previous clinical studies~\citep{Chansangpetch2018,Ang2018,Fu2019_AJO}.

From Table~\ref{tab_team_classification}, we found that most teams built their networks based on ResNet~\citep{ResNet2016} or SE-Net~\citep{SENET}. This demonstrates that basic deep networks have adequate ability to distinguish the angle closure. The top-two teams utilized advanced deep networks, \eg, Res2Net~\citep{gao2019res2net} and EfficientNet~\citep{Tan2019}, and got better performances. However, due to the limited amount of training data, the deep networks tended to suffer from overfitting. Combining multiple models or multi-scale features is a way to prevent this. Table~\ref{tab_team_classification} shows that six out of the eight teams utilized ensembling to improve the generalization performance.

\subsubsection{Clinical Discussion}
In clinical practice, the localization of the SS is the basic step to quantitatively evaluate the ACA. Therefore, we set up an independent task to automatically annotate the SS, and then calculate the AOD accordingly. Compared to the ground truth, the deep learning algorithms had an average deviation in SS localization of around 10$\mu$m. Further improvements are thus needed before they can be used in clinics. In the task of angle closure classification, all the algorithms achieved ideal performances, with nearly 100\% accuracy rate. This is understandable since the cases included in the AGE challenge tend to have common ACA morphology and no special structures such as plateau iris. 
Although our AGE challenge, composed of 4800 images, is currently the largest public dataset, we are still unable to predict if the algorithms would maintain good performance in a real-world setting, as the ACA morphologies are even more complex in the general population. This is a very promising start but there is still a long way to go.
Another potential limitation of our study is that the AS-OCT images were only taken using a Casia SS-1000 OCT device. This could possibly have a negative effect on the quality and performance when the algorithms are applied to images from other AS-OCT acquisition devices. In a future challenge, it would be of value to add more AS-OCT modalities from different-stage angle closure patients and train the algorithms for diagnosis.

\section{Conclusion}

In this paper, we summarized the methods and results of the AGE challenge. We compared the performances of the eight teams that participated in the onsite challenge at MICCAI 2019. Artificial intelligence techniques were shown to be promising for helping clinicians to reliably and rapidly identify SS points. Further, using deep learning methods to discriminate moderate or advanced angle closure from open angle also demonstrated encouraging results. 
 
In summary, the AGE challenge is the first open AS-OCT dataset focused on scleral spur localization and angle closure classification. The data and evaluation framework are publicly accessible through the Grand Challenges website at \curl{https://age.grand-challenge.org}.  Future participants are welcome to submit their results on the challenge website and use it for benchmarking their methods. The website will remain permanently available for submissions, to encourage future developments in the field. We expect that the unique AGE challenge will be beneficial to both early-stage and senior researchers in related fields.

\bibliographystyle{model2-names.bst}\biboptions{authoryear}
\bibliography{AGE_Challenge}

\end{document}